\documentclass{llncs}
\usepackage{graphicx} 
\usepackage{subfigure} 
  
\usepackage{algorithm}
\usepackage{algorithmic}
\usepackage{hyperref}

\newcommand{\algorithmicfunctionend}{\textbf{end function}\ }
\usepackage{amssymb}
\usepackage{amsmath}
\usepackage{times}

\newcommand{\minmax}{{MinMax}}
\newcommand{\JS}{{JS}}
\newcommand{\random}{{Random}}
\newcommand{\explEM}{Exploratory EM}
\newcommand{\explKM}{Explore-KMeans}
\newcommand{\semisupKM}{Semisup-KMeans}
\newcommand{\semisupEM}{SemisupEM}
\newcommand{\km}{K-Means}
\newcommand{\eat}[1]{}

\begin{document} 
\title{Exploratory Learning}
\date{}

\author{Bhavana Dalvi \and William W. Cohen \and Jamie Callan}
\institute{School of Computer Science, \\ Carnegie Mellon University, \\ Pittsburgh, PA 15213 \\  
\email{\{bbd, wcohen, callan\}@cs.cmu.edu}}

\maketitle

\begin{abstract}
In multiclass semi-supervised learning (SSL), it is sometimes the case
that the number of classes present in the data is not known, and hence
no labeled examples are provided for some classes.  In this paper we
present variants of well-known semi-supervised multiclass learning
methods that are robust when the data contains an unknown number of
classes.  In particular, we present an ``exploratory'' extension of
expectation-maximization (EM) that explores different numbers of
classes while learning.  ``Exploratory'' SSL greatly improves
performance on three datasets in terms of F1
on the classes \emph{with} seed examples---i.e., the classes
which are expected to be in the data.  Our \explEM{} algorithm
also outperforms a SSL method based non-parametric Bayesian
clustering.
\end{abstract}

\section{Introduction}
\vspace{-0.1in}
In multiclass semi-supervised learning (SSL), it is sometimes the case
that the number of classes present in the data is not known.  For
example, consider the task of classifying noun phrases into a large
hierarchical set of categories such as ``person'', ``organization'',
``sports team'', etc., as is done in broad-domain information
extraction systems (e.g., \cite{Carlson:wsdm2010}).  A
sufficiently large corpus will certainly contain some unanticipated
natural clusters---e.g., kinds of musical scales, or types of dental
procedures.  Hence, it is unrealistic to assume some examples have
been provided for each class: a more plausible assumption is that an
unknown number of classes exist in the data, and that labeled examples
have been provided for some subset of these classes.

This raises the natural question: how robust are existing SSL methods
to unanticipated classes?  As we will show experimentally below, SSL
methods can perform quite poorly in this setting: the instances of the
unanticipated classes might be forced into one or more of the expected
classes, leading to a cascade of errors in class parameters, and then
to class assignments to other unlabeled examples.  To address this
problem, we present an ``exploratory'' extension of
expectation-maximization (EM) which explores different numbers of
classes while learning.

More precisely, in a traditional SSL task, the learner assumes a fixed
set of classes $C_1, C_2 ,\ldots C_k$, and the task is to construct a
$k$-class classifier using labeled datapoints $X^l$ and unlabeled
datapoints $X^u$, where $X^l$ contains a (usually small) set of
``seed'' examples of each class.  In exploratory SSL, we assume the
same inputs, but allow the classifier to predict labels from the set
$C_1,\ldots,C_m$, where $m\geq{}k$: in other words, every example $x$
may be predicted to be in either a known class $C_i \in {C_1\ldots
  C_k}$, or an unknown class $C_i \in {C_{k+1}\ldots C_m}$.

We will show that exploratory SSL can greatly improve performance on
noun-phrase classification tasks and document classification tasks,
for several well-known SSL methods.  
E.g. Figure~\ref{fig:DelSp20NGConfusion} (b) top row shows, the
confusion matrices for a traditional SSL method on a 20-class problem 
 at the end of iteration 1 and 15, when the 
algorithm is presented with seeds for 6 of the classes.  
Here, red indicates overlap between classes, and
dark blue indicates no overlap. So we see that many of the seed classes are getting
confused with the unknown classes at the end of 15 iterations of 
SSL showing semantic drift. With the same inputs, our novel
``exploratory'' EM algorithm performs quite well 
(Figure \ref{fig:DelSp20NGConfusion} (b) bottom row); i.e.
it introduces additional clusters and at the end of 15 iterations 
improves F1 on classes for which seed examples were provided.\\

\noindent \textbf{Contributions.}
We  focus on the novel problem of dealing with learning when
only fraction of classes are known upfront, and there are unknown classes hidden in the data.
We propose a variant of the EM algorithm where new classes can be introduced in
each EM iteration. We discuss the connections of this algorithm to the structural EM algorithm. 
Next we propose two heuristic criteria 
for predicting when to create new class during an EM iteration,
and show that these two criteria work well on three publicly available datasets.
Further we evaluate third criterion, that introduces classes
uniformly at random and show that
our proposed heuristics are more effective than this baseline.
Experimentally, \explEM{} outperforms a semi-supervised variant of  
 non-parametric Bayesian clustering (Gibbs sampling with Chinese Restaurant Process)---a technique which
also ``explores'' different numbers of classes while learning. 
We also compare our method against a semi-supervised EM 
method with $m$ extra classes (trying different values of $m$).

In this paper, \explEM{} is instantiated to produce exploratory
versions of three well-known SSL methods: semi-supervised Naive Bayes,
seeded \km, and a seeded version of EM using a von Mises-Fisher
distribution \cite{Banerjee:jmlr2005}.  Our experiments focus on
improving accuracy on the classes that do have seed
examples---i.e., the classes which are expected to be in the data. \\

\noindent \textbf{Outline.} In Section \ref{sect:methods}, 
we first introduce an exploratory version of EM, and then
discuss several instantiations of it, based on different models for
the classifiers (mixtures of multinomials, \km, and mixtures of
von Mises-Fisher distributions) and different approaches to
introducing new classes.  We then compare against an alternative exploratory
SSL approach, namely Gibbs sampling with Chinese restaurant
process \cite{griffiths2004hierarchical}.  
Section \ref{sect:expt} presents experimental results, 
followed by related work and conclusions.

\section{Exploratory SSL Methods} \label{sect:methods}
\vspace{-0.1in}
\subsection{A Generic Exploratory Learner}

Many common approaches to SSL are based on EM.  In a typical EM
setting, the M-step finds the best parameters $\theta$ to fit the
data, $X^l \cup X^u$, and the E-step probabilistically labels the
unknown points with a distribution over the known classes $C_1, C_2
,\ldots C_k$.  In some variants of EM, including the ones we consider
here, a ``hard'' assignment is made to classes instead, an approach
named \emph{classification EM}
\cite{celeux1992classification}.  Our exploratory version of EM
differs in that it can introduce new classes ${C_{k+1}\ldots C_m}$ during
the E-step.
\begin{algorithm*}[htb]
\small   
\caption{EM algorithm for exploratory learning with model selection}
\label{Algo:ExploratoryEM}
\begin{algorithmic}[1]
\STATE \algorithmicfunction \ \textbf{\explEM\ } ($X^l$, $Y^l$, $X^u$, \{$C_1 \ldots C_k$\}): 
$\{C_{k+1}\ldots C_m\}$, $\theta^m$, $Y^u$
\STATE \textbf{Input}: $X^l$ labeled data points;  $Y^l$ labels for datapoints $X^l$;
               $X^u$ unlabeled datapoints (same feature space as $X^l$);
	       \{$C_1 \ldots C_k$\} set of known classes to which $x$'s belong.
\STATE \textbf{Output:}  $\{C_{k+1}\ldots C_m\}$ {newly-discovered classes}; 
        $\{\theta^1,\ldots,\theta^m\}$ {parameters for all $m$ classes}; 
        $Y^u$ labels for unlabeled data points $X^u$ \\
        \item[] \COMMENT{{Initialize model parameters using labeled data}}
	\STATE $\theta^1_0,\ldots,\theta^k_0 = argmax_{\theta} L(X^l, Y^l | \theta^k)$
	\STATE $i$ is \# new classes ; $i = 0$; \textit{CanAddClasses} = $true$
        \WHILE {data likelihood not converged AND \#classes not converged}
	   \item[] \COMMENT {{E step:} (Iteration $t$) Make predictions for the unlabeled data-points}
           \STATE $i_{old} = i$; Compute baseline log-likelihood \textit{BaselineLL} $= log P(X | \theta^{1}_t,\ldots,\theta^{k+i_{old}}_t)$
	   \FOR {$x \in X^u$}
		\STATE Predict $P(C_j|x,\theta^{1}_t,\ldots,\theta^{k+i}_t)$ for all labels $1\leq{}j\leq{k+i}$ \label{step:probs}

		\IF {nearlyUniform($P(C_1|x),\ldots,P(C_{k+i}|x)$) AND \textit{CanAddClasses}}

                  \STATE Increment $i$; Let $C_{k+i}$ be the new class. 
	          \STATE Label $x$ with $C_{k+i}$ in $Y^u$, and compute parameters $\theta^{k+i}_t$ for the new class.

		\ELSE
		  \STATE Assign $x$ to $(argmax_{C_j} P(C_j | x))$ in $Y^u$ where $1\leq{}j\leq{k+i}$
		\ENDIF
	   \ENDFOR
	   \STATE $i_{new} = i$; Compute ExploreEM loglikelihood \textit{ExploreLL} $= log P(X | \theta^{1}_t,\ldots,\theta^{k+i_{new}}_t)$
	   \item[] \COMMENT {{M step :} Recompute model parameters using current assignments for $X^u$}
           \IF {Penalized data likelihood is better for exploratory model than baseline model}
              \item[] \COMMENT {Adopt the new model with $k+i_{new}$ classes}
	      \STATE $\theta^{k+i_{new}}_{t+1} = argmax_{\theta} L(X^l, Y^l, X^u, Y^u_t | \theta^{k+i_{new}})$
	   \ELSE
	      \item[] \COMMENT {Keep the old model with $k+i_{old}$ classes}		      
	      \STATE $\theta^{k+i_{old}}_{t+1} = argmax_{\theta} L(X^l, Y^l, X^u, Y^u_t | \theta^{k+i_{old}})$
	      \STATE \textit{CanAddClasses} = $false$
	   \ENDIF
        \ENDWHILE
\STATE \algorithmicfunctionend
\end{algorithmic}
\end{algorithm*}

Algorithm \ref{Algo:ExploratoryEM} presents a generic \explEM{}
algorithm (without specifying the model being used).  There are two
main differences between the algorithm and standard classification-EM
approaches to SSL.  
First, in the E step, each of
the unlabeled datapoint $x$ is either assigned to one of the existing
classes, or to a newly-created class. We will discuss the
``near\-Uniform'' routine below, but the intuition we use is that a
new class should be introduced to hold $x$ when the probability of
$x$ belonging to existing classes is close to uniform.  This suggests
that $x$ is not a good fit to any existing classes, and that adding
$x$ to any existing class will lower the total data likelihood
substantially.
Second, in the M-step of iteration $t$, we choose either the model 
proposed by Exploratory EM method that might have more number 
of classes than previous iteration $t-1$ or the baseline
version with same number of classes as iteration $t-1$.
This choice is based on whether exploratory model satisfies a model 
selection criterion in terms of increased data likelihood and 
model complexity. If the algorithm decides that baseline model 
is better than exploratory model in iteration $t$, 
then from iteration $t+1$ onwards the algorithm won't introduce any new classes.

\vspace{-0.15in}
\subsection{Discussion}
\vspace{-0.1in}
Friedman \cite{friedman:jcb2002} proposed the Structural EM algorithm
that combines the standard EM algorithm, which optimizes parameters, with structure
search for model selection. This algorithm learns
networks based on penalized likelihood scores, in the presence of missing data.
In each iteration it evaluates multiple models based on the expected scores of models with missing data,
and selects the model with best expected score. 
This algorithm converges at local maxima for penalized log likelihood 
(the score includes penalty for increased model complexity).

Similar to Structural EM, in each iteration of Algorithm \ref{Algo:ExploratoryEM}, 
we evaluate two models, one with and one without adding extra classes. 
These two models are scored using a model selection criterion like AIC or BIC,
and the model with best penalized data likelihood score is selected in each iteration.
Further when the model selection criterion fails, the algorithm
reverts to standard semi-supervised EM algorithm.
Say this model switch happens at iteration $t_{switch}$, then 
from iteration 1 to $t_{switch}$, Algorithm \ref{Algo:ExploratoryEM} acts like the
structural EM algorithm \cite{friedman:jcb2002}.
From iteration $t_{switch}+1$ till the data likelihood converges, the algorithm acts as semi-supervised EM
algorithm. 

Next let us discuss the applicability of this algorithm for clustering
as well as classification tasks.
Notice that Algorithm~\ref{Algo:ExploratoryEM} reverts to an
unsupervised clustering method if $X_l$ is empty, and reverts to a
supervised generative learner if $X_u$ is empty.  Likewise, if no new
classes are generated, then it behaves as a multiclass SSL method; for
instance, if the classes are well-separated and $X_l$ contains enough
labels for every class to approximate these classes, then it is
unlikely that the criterion of nearly-uniform class probabilities will
be met, and the algorithm reverts to SSL.  Henceforth we will use the
terms ``class'' and ``cluster'' interchangeably.

\vspace{-0.15in}
\subsection{Model Selection}
\vspace{-0.1in}
For model penalties we tried multiple well known criteria like BIC, AIC and AICc.
Burnham and Anderson \cite{burnham:SMR2004} have 
experimented with AIC criteria and proposed AICc for datasets where,
the number of datapoints is less than 40 times number of features. 
The formulae for scoring a model using each of the three criteria that we tried are:
\begin{eqnarray}
BIC(g) &=& -2*L(g) + v*ln(n) \\
AIC(g) &=& -2*L(g) + 2*v  \\
AICc(g) &=& AIC(g) + 2*v*(v+1)/(n-v-1) 
\end{eqnarray}
where $g$ is the model being evaluated, $L(g)$ is the log-likelihood of the data given $g$, 
$v$ is the number of free parameters of the model and $n$ is the number of data-points.\\
While comparing two models, a lower value is preferred.
The extended Akaike information criterion (AICc) suited best for our experiments
since our datasets have large number of features and
small number of data points. With AICc criterion, the objective function that Algorithm \ref{Algo:ExploratoryEM} optimizes is:
\begin{eqnarray}
\max_{\substack{m, \{\theta^{1} \ldots \theta^{m}\}}, m\geq k} \{\text{Log Data Likelihood} - \text{Model penalty}\} \nonumber \\
\text{i.e., }\max_{\substack{m, \{\theta^{1} \ldots \theta^{m}\}, m \geq k}} \{ \log P(X | \theta^{1},\ldots,\theta^{m})\} - \{ v + (v*(v+1)/(n-v-1))\} 
\end{eqnarray}
Here, $k$ is the number of seed classes given as input to the algorithm and 
$m$ is the number of classes in the resultant model ($m \geq k$).

\subsection{Exploratory versions of well-known SSL methods}
In this section we will consider various SSL techniques, and propose
exploratory extensions of these algorithms.

\vspace{-0.15in}
\subsubsection{Semi-Supervised Naive Bayes}
Nigam et al. \cite{nigam:mlj2000} proposed an EM-based semi-supervised
version of multinomial Naive Bayes.  In this model $P(C_j | x) \propto
P(x | C_j) * P(C_j)$, for each unlabeled point $x$.  The probability
$P(x | C_j)$ is estimated by treating each feature in $x$ as an
independent draw from a class-specific multinomial.  In document
classification, the features are word occurrences, and the number of
outcomes of the multinomial is the vocabulary size.  

This method can be naturally used as an instance of \explEM, using the
multinomial model to compute $P(C_j|x)$ in Line~\ref{step:probs}.  The
M step is also trivial, requiring only estimates of $P(w|C_j)$ for
each word/feature $w$. 

\vspace{-0.15in}
\subsubsection{Seeded \km}
It has often been observed that \km{} and EM are algorithmically
similar.  Basu and Mooney \cite{Basu:icml2002} proposed a seeded
version of \km, which is very analogous to Nigam et al's
semi-supervised Naive Bayes, as another technique for semi-supervised
learning.  Seeded \km{} takes as input a number of clusters, and
seed examples for each cluster.  The seeds are used to define an
initial set of cluster centroids, and then the algorithm iterates
between an ``E step'' (assigning unlabeled points to the closest
centroid) and an ``M step'' (recomputing the centroids).

In the seeded \km{} instance of \explEM, we again define $P(C_j |
x) \propto P(x | C_j) * P(C_j)$, but define 
$P(x | C_j) = x \cdot C_j$, i.e., the inner product of a vector representing $x$
and a vector representing the centroid of cluster $j$.  Specifically,
$x$ and $C_j$ both are represented as $L_1$ normalized
TFIDF feature vectors.  The centroid of a new cluster is initialized
with smoothed counts from $x$. 
In the ``M step'', we recompute the centroids of clusters in the usual
way.

\vspace{-0.15in}
\subsubsection{Seeded Von-Mises Fisher}
The connection between \km{} and EM is explicated by Banerjee et
al. \cite{Banerjee:jmlr2005}, who described an EM algorithm that is directly
inspired by \km{} and TFIDF-based representations.  In particular,
they describe generative cluster models based on the von Mises-Fisher
(vMF) distribution, which describes data distributed on the unit
hypersphere.  Here we consider the ``hard-EM'' algorithm proposed by
Banerjee et al, and use it in the seeded (semi-supervised) setting
proposed by Basu et al. \cite{Basu:icml2002}.  This natural extension of Banerjee et
al\cite{Banerjee:jmlr2005}'s work can be easily extended to our exploratory setting.

As in seeded \km, the parameters of vMF distribution are
initialized using the seed examples for each known cluster.  In each
iteration, we compute the probability of $C_j$ given data point $x$,
using vMF distribution, and then assign $x$ to the cluster for which
this probability is maximized.  The parameters of the vMF distribution
for each cluster are then recomputed in the M step.  For this method,
we use a TFIDF-based $L_2$ normalized vectors, which lie on the unit
hypersphere.

Seeded vMF and seeded \km{} are closely related---in particular,
seeded vMF can be viewed as a more probabilistically principled
version of seeded \km.  Both methods allow use of TFIDF-based
representations, which are often preferable to unigram representations
for text: for instance, it is well-known that unigram representations
often produce very inaccurate probability estimates.

\vspace{-0.1in}
\subsection{Strategies for inducing new clusters/classes}
\vspace{-0.1in}
\begin{algorithm}[tb]
\small   
\caption{\JS\ criterion for new class creation}
\label{Algo:KLDivergenceClassCriterion}
\begin{algorithmic}[1]
\STATE \algorithmicfunction \ \textbf{JSCriterion}($[P(C_1 | x)\ \ldots \ P(C_k | x)]$): 
\STATE \textbf{Input}: $[P(C_1 | x) \ldots P(C_k | x)]$ probability distribution of existing classes for a data point $x$
\STATE \textbf{Output:}  \textit{decision} : \textbf{true} iff new class needs to be created
 	\STATE $u$ = [$1/k$ $\ldots$ $1/k$]  \COMMENT{i.e., the uniform distribution with current number of classes = $k$}
        \STATE \textit{decision} = false
 	\IF {Jensen-Shannon-Divergence($u$, $P(C_j|x)$) $<$ $\frac{1}{k}$}
 	  \STATE \textit{decision} = true
 	\ENDIF	
\STATE \algorithmicfunctionend
\end{algorithmic}
\end{algorithm}

\begin{algorithm}[tb]
\small   
\caption{\minmax\  criterion for new class creation}
\label{Algo:heuristicClassCriterion}
\begin{algorithmic}[1]
\STATE \algorithmicfunction \ \textbf{MinMaxCriterion}($[P(C_1 | x)\ \ldots \ P(C_k | x)]$): 
\STATE \textbf{Input}: $[P(C_1 | x) \ldots P(C_k | x)]$ probability distribution of existing classes for a data point $x$
\STATE \textbf{Output:}  \textit{decision} : \textbf{true} iff new class needs to be created
	\STATE $k$ is the current number of classes
 	\STATE $maxProb$ = max($P(C_j | x)$); $minProb$ = min($P(C_j | x)$)
 	\IF {$\frac{maxProb}{minProb} < 2$}
 	  \STATE \textit{decision} = true
 	\ENDIF	
\STATE \algorithmicfunctionend
\end{algorithmic}
\end{algorithm}
In this section we will formally describe some possible strategies for
introducing new classes in the E step of the algorithm.  They are
presented in detail in Algorithms
\ref{Algo:KLDivergenceClassCriterion} and
\ref{Algo:heuristicClassCriterion}, and each of these is a possible
implementation of the ``near\-Uniform'' subroutine of
Algorithm~\ref{Algo:ExploratoryEM}.
As noted above, the intuition is that new classes should be
introduced to hold $x$ when the probabilities of $x$ belonging to
existing classes are close to uniform.  In the \JS{} criterion,
we require that Jensen-Shanon divergence\footnote{The Jensen-Shannon
  divergence between $p$ and $q$ is the average Kullback-Leiber
  divergence of $p$ and $q$ to $a$, the average of $p$ and $q$, i.e.,
  $\frac{1}{2} (KL(p||a + KL(q||a))$.} between the posterior class
distribution for $x$ to the uniform distribution
be less than $\frac{1}{k}$.  The \minmax{} criterion
is a somewhat simpler approximation to this intuition: a new cluster
is introduced if the maximum probability is no more than twice the
minimum probability. 

\subsection{Baseline Methods}
\vspace{-0.1in}
Next, we will take a look at various baseline methods that we implemented to 
measure the effectiveness of our proposed approach.

\subsubsection{Random new class creation criterion:}
To measure the effectiveness of criteria proposed in 
Algorithms \ref{Algo:KLDivergenceClassCriterion} and \ref{Algo:heuristicClassCriterion}, we experimented 
with a random baseline criterion, that returns ``true'' uniformly at random 
with probability equal to that of \minmax{} or \JS{} criterion returning true
for the same dataset. This is referred to as \random{} criterion below.

\begin{algorithm}[tb]
\small   
\caption{Exploratory Gibbs Sampling with Chinese Restaurant Process}
\label{Algo:ExploreGibbsCRP}
\begin{algorithmic}[1]
\STATE \algorithmicfunction \ \textbf{GibbsCRP} ($X^l$, $Y^l$, $X^u$, \{$C_1 \ldots C_k$\}) : 
${C_{k+1}\ldots C_m}$, $Y^u$
\STATE \textbf{Input}: $X^l$ labeled data points; 
                       $Y^l$ labels of $X^l$; 
                       $X^u$ unlabeled data points; \\
                       \{$C_1 \ldots C_k$\} set of known classes $x$'s belong to; 
                       $P_{new}$ probability of creating a new class.
\STATE \textbf{Output:} ${C_{k+1}\ldots C_m}$ {newly-discovered classes}; $Y^u$ labels for $X^u$ \\
 	\FOR {$x$ in $X^u$}
 	  \STATE Save a random class from \{$C_1 \ldots C_k$\} for $x$ in $Y^u$ 
 	\ENDFOR
	\STATE Set $m=k$
        \FOR {$t$ : 1 to $numEpochs$}
	    \FOR {$x_i$ in $X^u$}
		\STATE Let $y_i$'s be $x_i$'s label in epoch $t-1$
		\STATE predict $P(C_j|x_i,Y^l\cup{}Y^u-\{y_i\})$
                   ~~~for all labels $1\leq{}j\leq{}m$
                \STATE $y_i',m'$ = 
                    CRPPick($P_{new}$, $P(C_1|x_i),\ldots,P(C_{m+1}|x_i)$)
                \STATE Save $y_i'$ as $x_i$'s label in epoch $t$
		\STATE $m=m'$
	    \ENDFOR
	\ENDFOR 
\STATE \algorithmicfunctionend
\end{algorithmic}
\end{algorithm}

\begin{algorithm}[htb]
\small   
\caption{Modified CRP criterion for new class creation}
\label{Algo:CRPClassCriterion}
\begin{algorithmic}[1]
\STATE \algorithmicfunction \ \textbf{ModCRPPick} ($P_{new}$, $P(C_1 | x),\ldots,P(C_{k+i} | x)$) : $y,i'$
\STATE \textbf{Input}: $P_{new}$ probability of creating new class; \\ 
         $P(C_1 | x), \ldots, P(C_{k+i} | x)$ probability of existing classes given $x$
\STATE \textbf{Output:}
      $y$ class for $x$; 
      $i'$ new number of classes
 	\STATE $u$ = [$1/{k+i}$ $\ldots$ $1/{k+i}$] \COMMENT{uniform distribution with $k+i$ classes}
        \STATE $d$ = Jensen-Shannon-Divergence($u$, $P(C_j | x)$)
        \STATE $q$ = $\frac{P_{new}}{((k+i) * d)}$ 
        \IF {a coin with bias $q$ is heads}
           \item[] \COMMENT{create a new class and assign to that}\\
           \STATE $y=k+i+1$ and $i'=i+1$
        \ELSE
           \item[] \COMMENT{assign to an existing class}
           \STATE $i'=i$ and $y$ = sample from distribution $[P(C_1 | x)\ \ldots \ P(C_k | x)]$
 	\ENDIF	
\STATE \algorithmicfunctionend
\end{algorithmic}
\end{algorithm}

\subsubsection{Semi-supervised EM with $m$ extra classes:} \label{sect:algo-overshoot-semisupEM}
\vspace{-0.1in}
One might argue that the goal of the \explEM{} algorithm can also be achieved by 
adding a random number of empty classes to the semi-supervised EM algorithm.
We compare our method against the best possible value of this baseline,
i.e. by choosing the number of classes that maximizes F1 on the seed classes.
Note that in practice, the test labels are not available, so this is the upper bound on
performance of this baseline. 
We compare our method with this upper bound in Section \ref{sect:expt}.
Our method is different from this baseline in two ways. 
First, it does not need the number of extra clusters as input.
Second, it seeds the extra clusters with those datapoints 
that are unlikely to belong to existing classes, as compared to
initializing them randomly.

\subsubsection{A seeded Gibbs sampler with CRP:} \label{sect:algo-crp}
The \explEM{} method is broadly similar to non-parametric
Bayesian methods, such as the Chinese Restaurant process (CRP)
\cite{griffiths2004hierarchical}.  CRP is often used in
non-parametric models (e.g., topic models) that are based on Gibbs
sampling, and indeed, since it is straightforward to replace EM with
Gibbs-sampling, one can use this approach to estimate the parameters of any of the
models considered here (i.e., multinomial Naive Bayes, \km,
and the von Mises-Fisher distribution).
Algorithm~\ref{Algo:ExploreGibbsCRP} presents a seeded version of a
Gibbs sampler based on this idea.  In brief,
Algorithm~\ref{Algo:ExploreGibbsCRP}, starts with a classifier trained
on the labeled data. Collapsed Gibbs sampling is then performed over
the latent labels of unlabeled data, incorporating the CRP into the
Gibbs sampling to introduce new classes. (In fact, we use block
sampling for these variables, to make the method more similar to the
EM variants.)

Note that this algorithm is naturally ``exploratory'', in our sense,
as it can produce a number of classes larger than the number of
classes for which seed labels exist.  However, unlike our exploratory
EM variants, the introduction of new classes is not driven by examples
that are ``hard to classify''---i.e., have nearly-uniform posterior
probability of membership in existing classes.  
In CRP method, the probability of creating a new class depends 
on the data point, but it does not explicitly favor cases
where the posterior over existing classes is nearly uniform.

To address this issue, we also implemented a variant of the seeded
Gibbs sampler with CRP, in which the examples with nearly-uniform
distributions are more likely to be assigned to new classes.  This variant
is shown in Algorithm~\ref{Algo:CRPClassCriterion}, which replaces the
routine CRP\-Pick in the Gibbs sampler---in brief, we simply scale
down the probability of creating a new class by the Jensen-Shannon
divergence of the posterior class distribution for $x$ to the uniform
distribution.  Hence the probability of creating new class explicitly
depends on how well the given data point fits in one of the
existing classes. An experimental comparison of our proposed method
with Gibbs sampling and CRP based baselines is shown in Section \ref{sect:expt:crp}.

\section{Experimental Results} \label{sect:expt}
\vspace{-0.1in}
We now seek to experimentally answer the questions raised in the
introduction.  How robust are existing SSL methods, if they are given
incorrect information about the number of classes present in the data,
and seeds for only some of these classes?  Do the exploratory versions
of the SSL methods perform better?  How does \explEM{} compare
with the existing ``exploratory'' method of Gibbs sampling with CRP?

We used three publicly available datasets for our experiments.  The
first is the widely-used 20-Newsgroups dataset \cite{ds:20newsgroup}.
We used the ``bydate'' dataset, which contains total of 18,774 text
documents, with vocabulary size of 61,188.  There are 20 non-overlapping
classes and the entire dataset is labeled.
The second dataset is the Delicious\_Sports dataset, published by
\cite{Dalvi:wsdm2012}.  This is an entity classification dataset,
which contains items extracted from 57K HTML tables in the sports
domain (from pages that had been tagged by the social bookmarking
system del.icio.us).  The features of an entity are ids for the HTML
table columns in which it appears.  This dataset contains 282 labeled entities
described by 721 features and 
26 non-overlapping classes (e.g., ``NFL teams'', ``Cricket teams'').
The third dataset is the Reuters-21578 dataset published 
by Cai et al. \cite{cai:icml2009}. This corpus originally contained 21,578 documents 
from 135 overlapping categories. Cai et al. discarded documents with multiple category labels,
resulting in 8,293 documents (vocabulary size=18,933) in 65 non-overlapping categories.


\subsection{\explEM\  vs. \semisupEM\ with few seed classes}

\begin{table*}[t]
\centering
\boldmath
\scalebox{0.95}{%
 \begin{tabular}{|l|l|c|l|l|l|l|} 
\hline
Dataset        & Algorithm           & \semisupEM\     &  \multicolumn{3}{|c|} {{\explEM\ }}                      & Best $m$ extra classes \\   
                                       \cline{4-6}   
(\#seed / \#total classes) & &                             & \minmax     & \JS & \random      &  \semisupEM\ \\
\hline
Delicious\_Sports       & KM  & 60.9    & 89.5 (30) $\blacktriangle$      & \textbf{90.6} (46) $\blacktriangle$     & 84.8 (55) $\blacktriangle$   &  69.4 (10) $\blacktriangle$ \\
(5/26)                  & NB  & 46.3    & 45.4 (06)                       & \textbf{88.4} (51) $\blacktriangle$     & 67.8 (38) $\blacktriangle$   &  65.8 (10) $\blacktriangle$ \\
                        & VMF & 64.3    & 72.8 (06) $\vartriangle$        & 63.0 (06)                      & 66.7 (06)                    &  \textbf{78.2} (09) $\blacktriangle$ \\
\hline
20-Newsgroups           & KM  & 44.9    & \textbf{57.4} (22) $\blacktriangle$      & 39.4 (99) $\blacktriangledown$ & 53.0 (22)  $\blacktriangle$  &  49.8 (11) $\blacktriangle$ \\
 (6/20)                 & NB  & 34.0    & 34.6 (07)                       & 34.0 (06)                      & 34.0 (06)                    &  \textbf{35.0} (07)                  \\
                        & VMF & 18.2    & 09.5 (09) $\blacktriangledown$  & 19.8 (06)                      & 18.2 (06)                    &  \textbf{20.3} (10) $\blacktriangle$ \\
\hline
Reuters                 & KM  & 8.9     & 12.0 (16)     $\vartriangle$    &\textbf{27.4} (100) $\blacktriangle$     & 13.7 (19)  $\blacktriangle$  &  16.3 (14) $\blacktriangle$ \\
       (10/65)          & NB  & 6.4     & 10.4 (10)                       & \textbf{18.5} (77)  $\blacktriangle$    & 10.6 (10)                    &  16.1 (15)                  \\
                        & VMF & 10.5    & 20.7 (11)      $\blacktriangle$ & \textbf{30.4} (62)    $\blacktriangle$  & 10.5 (10)                    &  20.6 (16)  $\vartriangle$  \\
\hline
 \end{tabular}
}
\caption{Comparison of \explEM\  w.r.t. \semisupEM\ for different datasets and class creation criteria.
For each exploratory method we report the macro avg. F1 over seed classes followed 
by avg number of clusters generated.
e.g. For 20-Newsgroups dataset, \explEM{} with \km{} and \minmax{} results in  57.4 F1 and generates 22 clusters on avg.
 $\blacktriangle$ (and $\vartriangle$) 
indicates that improvements are statistically significant w.r.t \semisupEM\ with 0.05 (and 0.1) significance level.}
\label{table:EMSummary}
\vspace{-0.15in}
\end{table*}

Table \ref{table:EMSummary} shows the performance of seeded \km,
seeded Naive Bayes, and seeded vMF using 5 different algorithms.  
For each dataset only a few of
the classes present in the data (5 for Delicious\_Sports, and 6
for 20-Newsgroups and 10 for Reuters), are given as seed classes to all the
algorithms.  Five percent datapoints were given as training data for each ``seeded'' class.  
The first method, shown in the column labeled \semisupEM{}, uses these
methods as conventional SSL learners.
The second method is \explEM{} with the simple \minmax{} new-class
introduction criterion, and the third is \explEM{} with
the \JS{} criterion.  Forth method is \explEM{} with
the \random{} criterion. The last one is upper bound on \semisupEM{} with $m$ extra classes.

ExploreEM performs hard clustering of the dataset i.e. each datapoint belongs to only one cluster. 
For all methods, for each cluster we assign a label that maximizes accuracy (i.e. majority label for the cluster). 
Thus using complete set of labels we can generate a single label per datapoint. 
Reported Avg. F1 value is computed by macro averaging F1 values of seed classes only.
Note that, for a given dataset, number of seed classes and training percentage per seed class
there are many ways to generate a train-test partition.
We report results using 10 random train-test partitions of each dataset. The same partitions are
used to run all the algorithms being compared and to compute the statistical significance of results.

\begin{figure*}[tb]
\begin{center}
   \includegraphics[scale =.32] {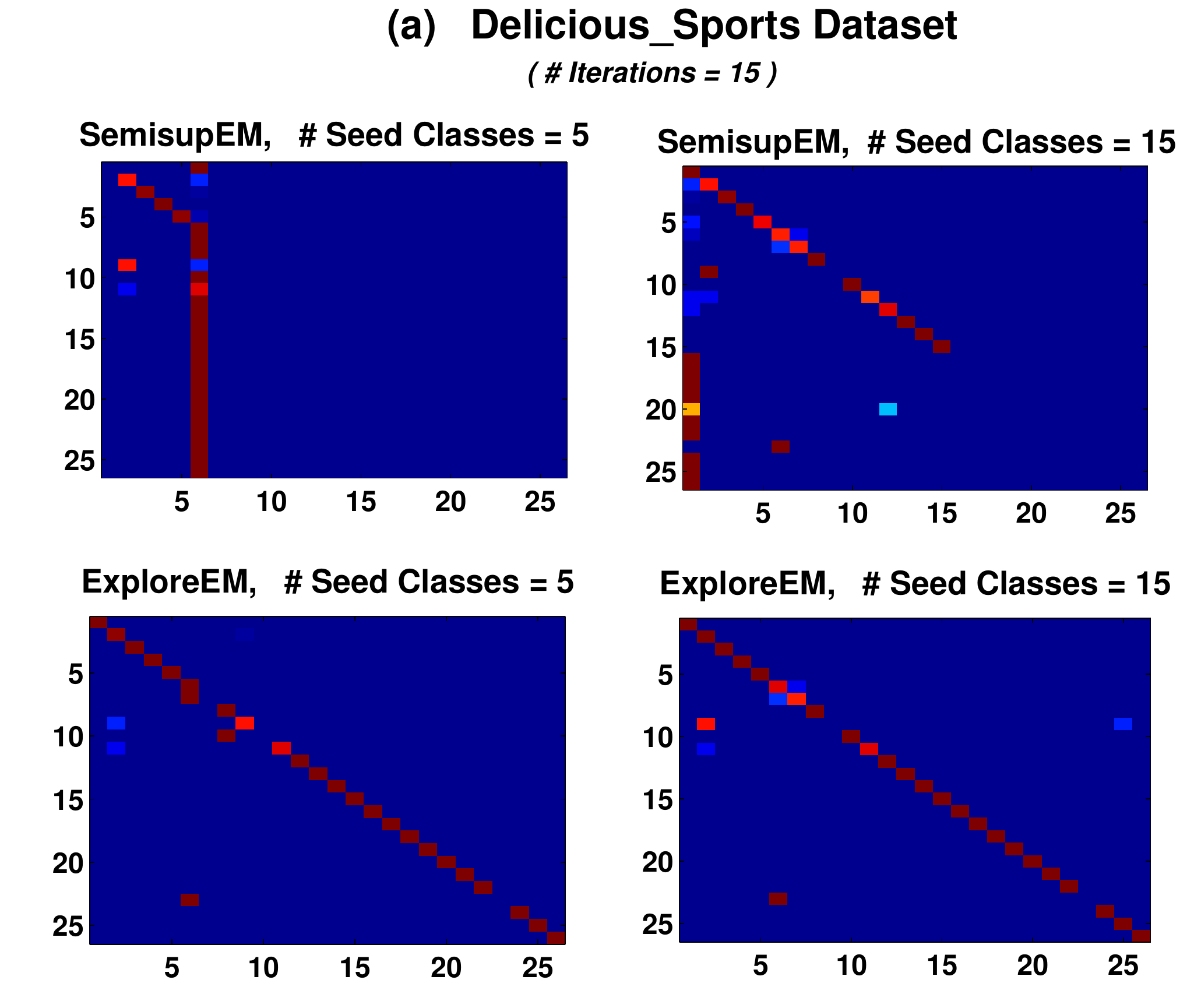}~~~~\includegraphics[scale =.32] {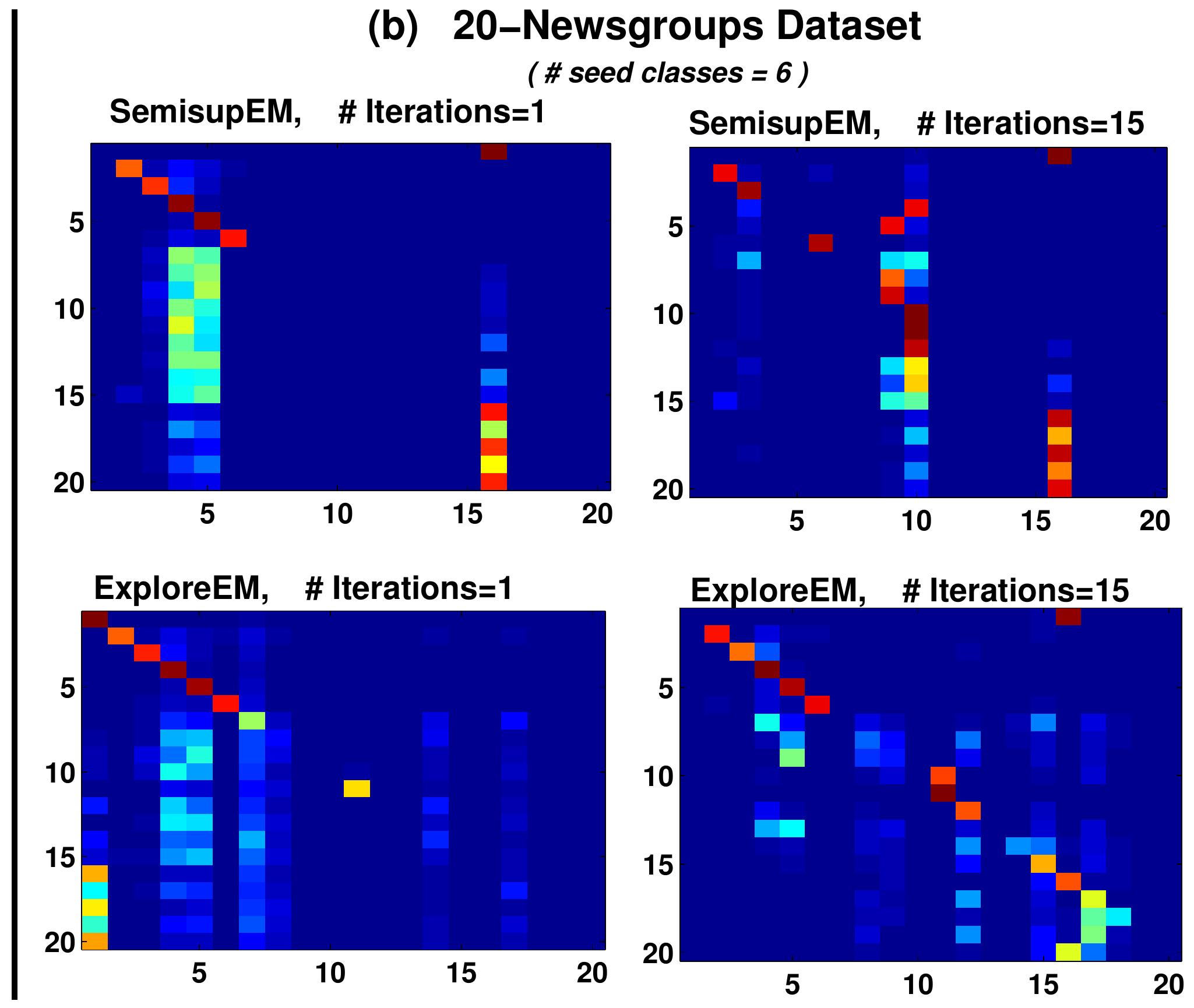}
\vspace{-0.15in}
\caption{(a) Confusion matrices, varying number of seed
  classes, for the Delicious\_Sports dataset.  (b) Confusion matrices, varying
  the number of EM iterations for the 20-Newsgroups dataset.  Each is
  using \explKM\ with the \minmax\ criterion.}
\label{fig:DelSp20NGConfusion}
\end{center}
\end{figure*}

We first consider the value of exploratory learning. With the
\JS{} criterion, the exploratory extension gives comparable or improved
performance on 8 out of 9 cases. In 5 out of 8 cases the gains are statistically significant.
With the simpler \minmax{} criterion, the exploratory extension 
results in performance improvements in 6 out of 8 cases,
and significantly reduces performance only in one case.  
The number of classes finally introduced by the
\minmax{} criterion is generally smaller than 
those introduced by \JS{} criterion.

For both SSL and exploratory systems, the seeded \km{}
method gives good results on all 3 datasets. In our MATLAB implementation, the running time of \explEM{} is longer, but not
unreasonably so: on average for 20-Newsgroups dataset  \semisupKM\ took 95 sec.
while \explKM\ took 195 sec. and for Reuters dataset, \semisupKM\ took 7 sec. while \explKM\ took
28 sec. 

We can also see that \random{} criterion shows significant improvements 
over the baseline \semisupEM{} method in 4 out of 9 cases.
While \explEM{} method with \minmax{} and \JS{} criterion shows 
significant improvements in 5 out of 9 cases.
In terms of magnitude of improvements, \JS{} is superior to \random{} criterion.

Next we compare \explEM{} with baseline named ``\semisupEM{} with $m$ extra classes''. 
The last column of Table \ref{table:EMSummary}
shows the best performance of this baseline by varying $m$ = \{0, 1, 2, 5, 10, 20, 40\},
and choosing that value of $m$ for which seed class F1 is maximum.
Since the ``best $m$ extra classes'' baseline is making use of the test labels
to pick right number of classes, it cannot be used in practice;
however \explEM{} methods produce comparable or better performance with 
this strong baseline. 

To better understand the qualitative behavior of our methods, we
conducted some further experiments with \semisupKM{} with the
\minmax{} criterion (which appears to be a reasonable baseline method.)
We constructed confusion matrices for the classification task, to
check how different methods perform on each dataset.\footnote{For
  purposes of visualization, introduced classes were aligned optimally
  with the true classes.}  Figure \ref{fig:DelSp20NGConfusion} (a) shows the
confusion matrices for \semisupEM{} (top row) and \explEM{} (bottom row)
 with five and fifteen seeded classes.  We
can see that \semisupEM\ with only five seed classes clearly confuses
the unexpected classes with the seed classes, while \explEM{} gives
better quality results.  Having seeds for more classes helps both
\semisupEM\ and \explEM, but \semisupEM{} still tends to confuse the
unexpected classes with the seed classes.
Figure \ref{fig:DelSp20NGConfusion} (b) shows similar results on the
20-Newsgroups dataset, but shows the confusion matrix after 1
iteration and after 15 iterations of EM.
It shows that \semisupEM{} after 15 iterations has made
limited progress in improving its classifier when compared to \explEM.

\begin{figure*}[tb]
\centering
   \includegraphics[scale =.45] {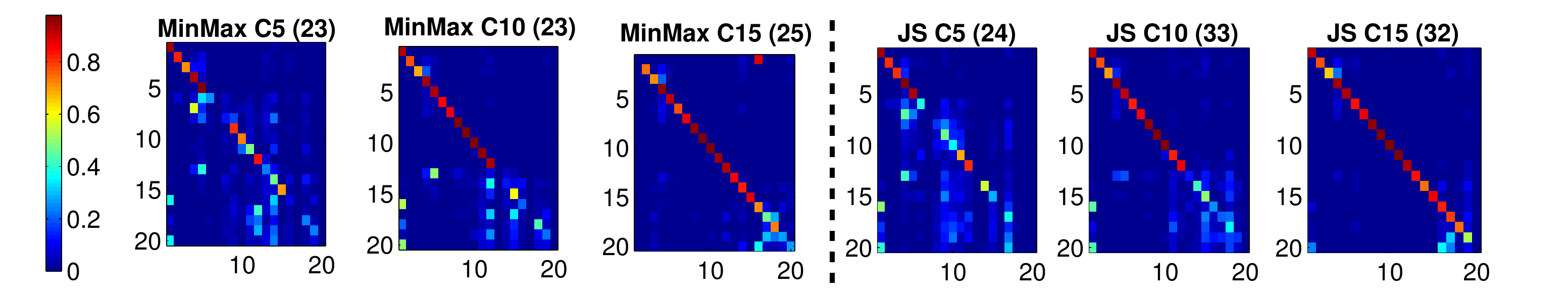}
\vspace{-0.25in}
\caption{20-Newsgroups dataset : Comparison of \minmax\ vs. \JS\ criterion for ExploreEM}
\label{fig:20NGConfusionHeuristic}
\end{figure*}

Finally, we compare the two class creation criteria, and show a
somewhat larger range of seeded classes, ranging from 5 to 15
(out of 20 actual classes).  In Figure~\ref{fig:20NGConfusionHeuristic}
each of the confusion-matrices is annotated with the strategy, the
number of seed classes and the number of classes produced. (E.g., plot
``MinMax-C5(23)'' describes \explKM\ with \minmax\ criterion and 
5 seed classes which produces 23 clusters.)  We can see that
\minmax\ criterion usually produces a more reasonable number of
clusters, closer to the ideal value of 20; however,
performance of the \JS{} method in terms of seed class accuracy
 is comparable to the \minmax{} method.

\begin{figure*}[tb]
\begin{minipage}[t]{0.45\textwidth}
\centerline{\includegraphics[width=\textwidth] {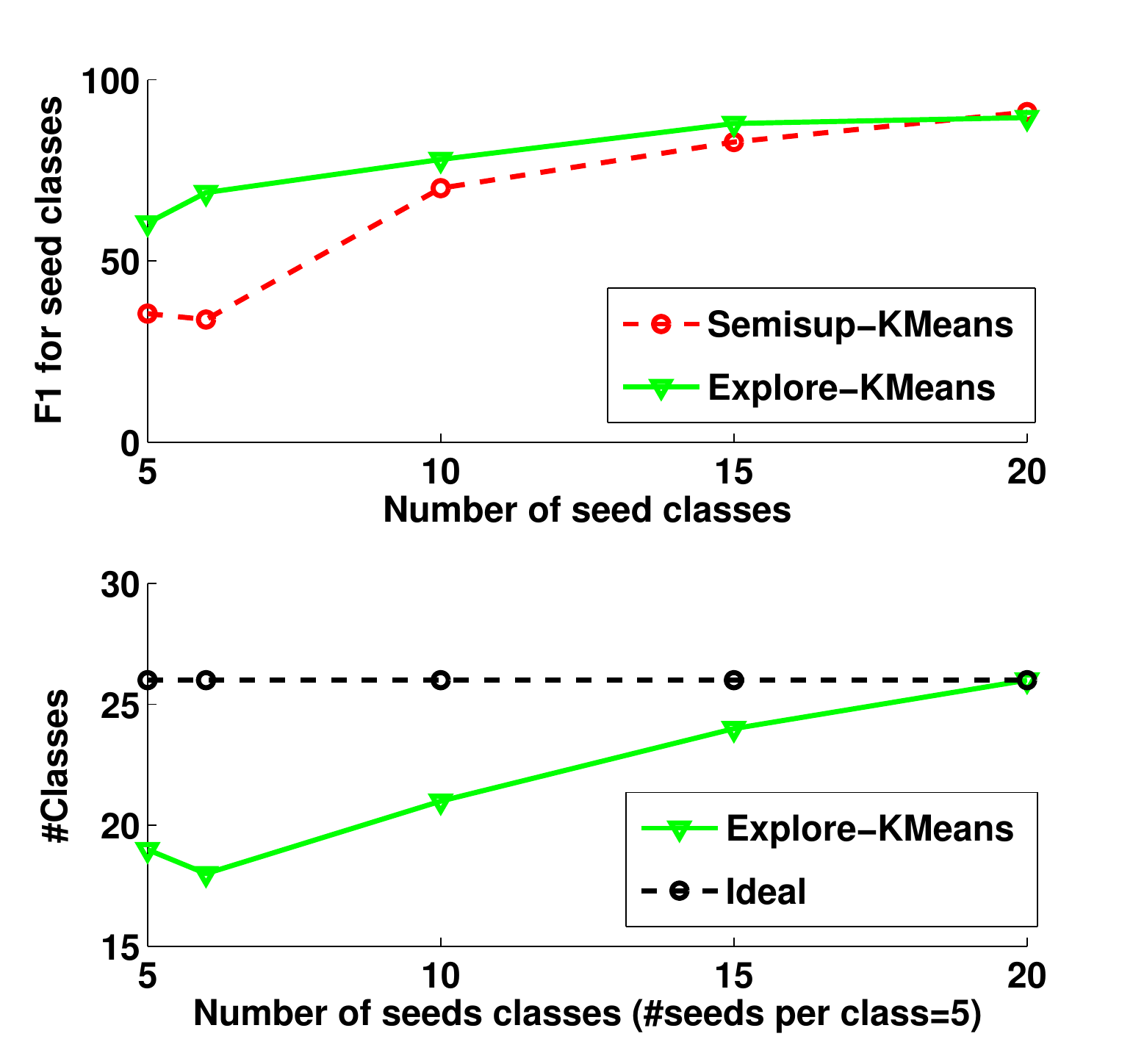}}
\vspace{-0.15in}
\caption{20-Newsgroups dataset: varying the number of seed classes (using the \minmax{} criterion).}
\label{fig:expt2-20NG-KM-PR-varyClass-varySeed}
\end{minipage}\hspace{0.3in}\begin{minipage}[t]{0.45\textwidth}
\centerline{\includegraphics[width=\textwidth] {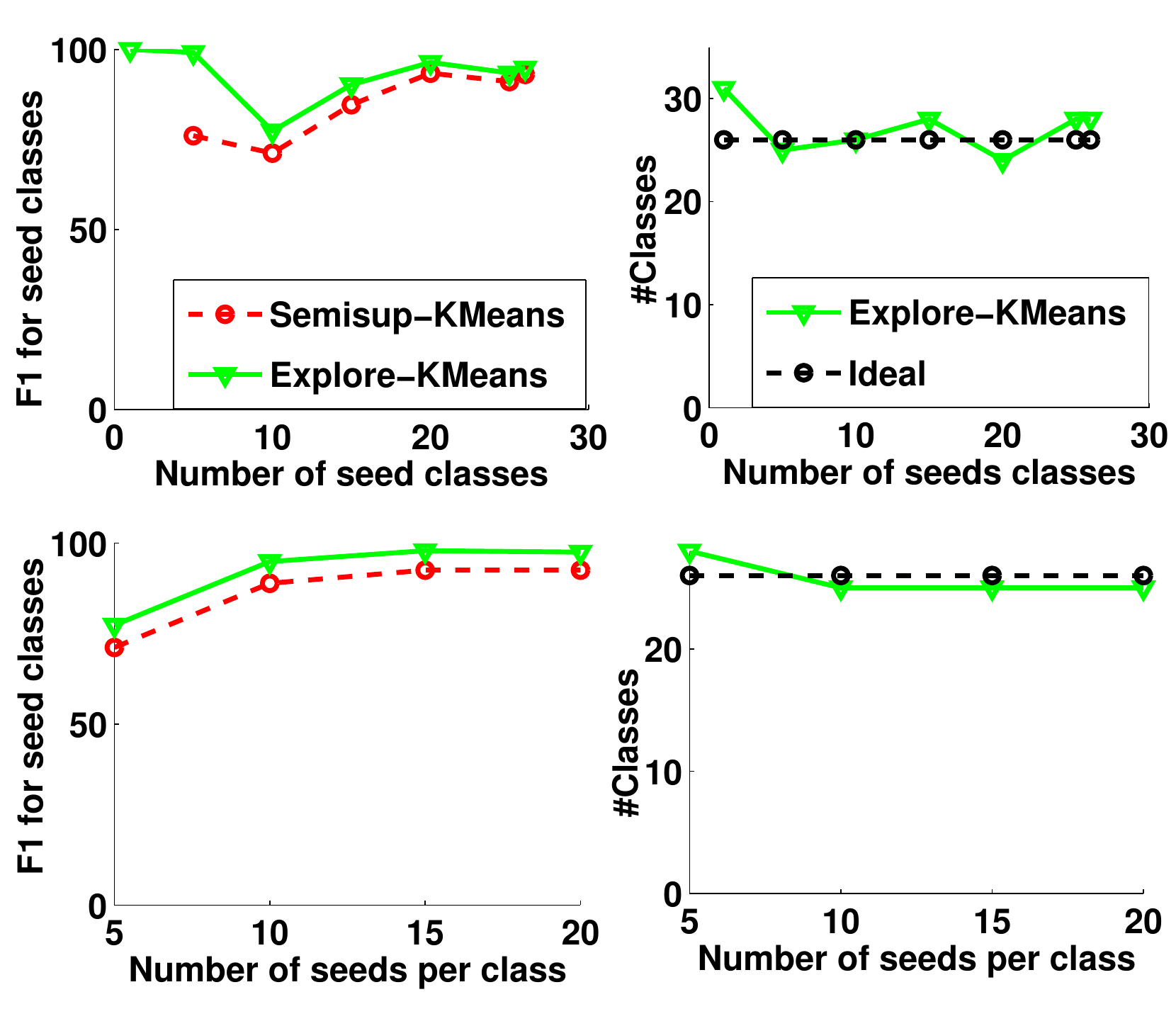}}
\vspace{-0.15in}
\caption{Delicious Sports dataset: Top, varying the number of seed classes (with five seeds per class).
Bottom, varying the number of seeds per class (with 10 seed classes).}
\label{fig:expt2-DelSp-KM-PR-varyClass-varySeed}
\end{minipage}
\end{figure*}

These trends are also shown quantitatively in Figure
\ref{fig:expt2-20NG-KM-PR-varyClass-varySeed}, which shows the result
of varying the number of seeded classes (with five seeds per class)
for \explKM{} and \semisupKM{}; the top shows the effect on F1, and
the bottom shows the effect on the number of classes produced (for
\explKM{} only).  Figure
\ref{fig:expt2-DelSp-KM-PR-varyClass-varySeed} shows a similar effect
on the Delicious\_Sports dataset: here we systematically vary the
number of seeded classes (using 5 seeds per seeded class, on the top),
and also vary the number of seeds per class (using 10 seeded
classes, on the bottom.)  The left-hand side compares the F1 for
\semisupKM{} and \explKM{}, and the right-hand side shows the
number of classes produced by \explKM{}.  For all parameter
settings, \explKM\ is better than or comparable to 
 \semisupKM\ in terms of F1 on seed classes. 

\subsection{Comparison with the Chinese Restaurant Process} \label{sect:expt:crp}
\vspace{-0.1in}
As discussed in Section \ref{sect:algo-crp}, a seeded version of the
Chinese Restaurant Process with Gibbs sampling (CRP−Gibbs) is an
alternative exploratory learning algorithm.  In this section we
compare the performance of CRP−Gibbs with \explKM\ and \semisupKM.  We
consider two versions of CRP-Gibbs, one using the standard CRP and one
using our proposed modified CRP criterion for new class creation
that is sensitive to the near-uniformity of instance's posterior class
distribution.  CRP-Gibbs uses the same instance representation as our
\km{} variants i.e. $L_1$ normalized TFIDF features.

\begin{figure*}[t]
\centering
   \includegraphics[scale =.3] {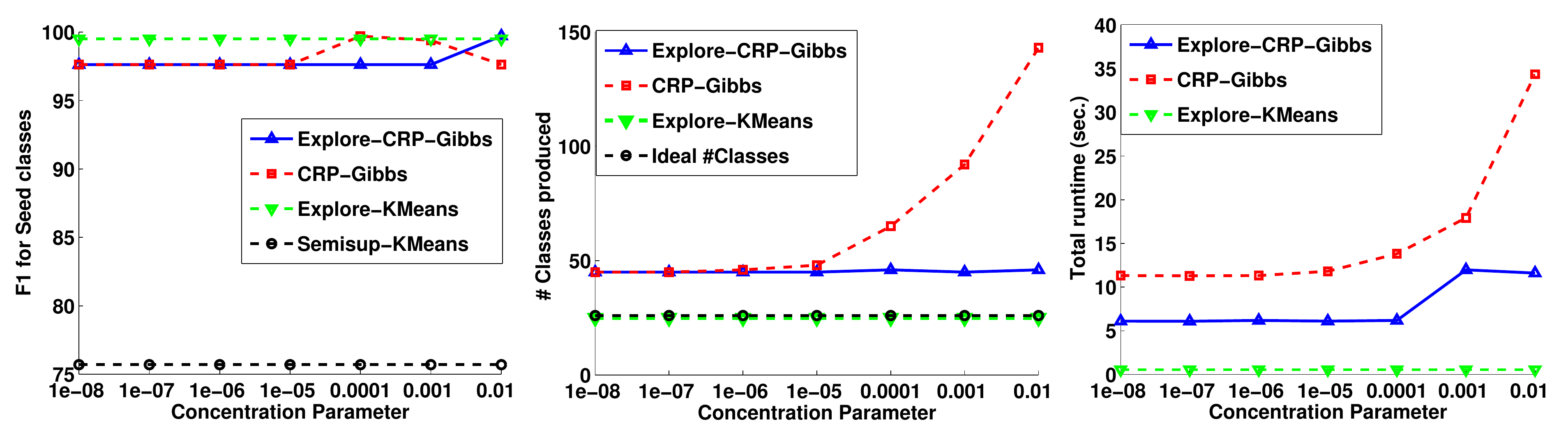}
\vspace{-0.15in}
\caption{Delicious\_Sports dataset: Varying the concentration parameter, with five seed classes.}
\label{fig:DelSpVaryPrior}
\end{figure*}

\begin{figure*}[htb]
\centering
   \includegraphics[scale =.33] {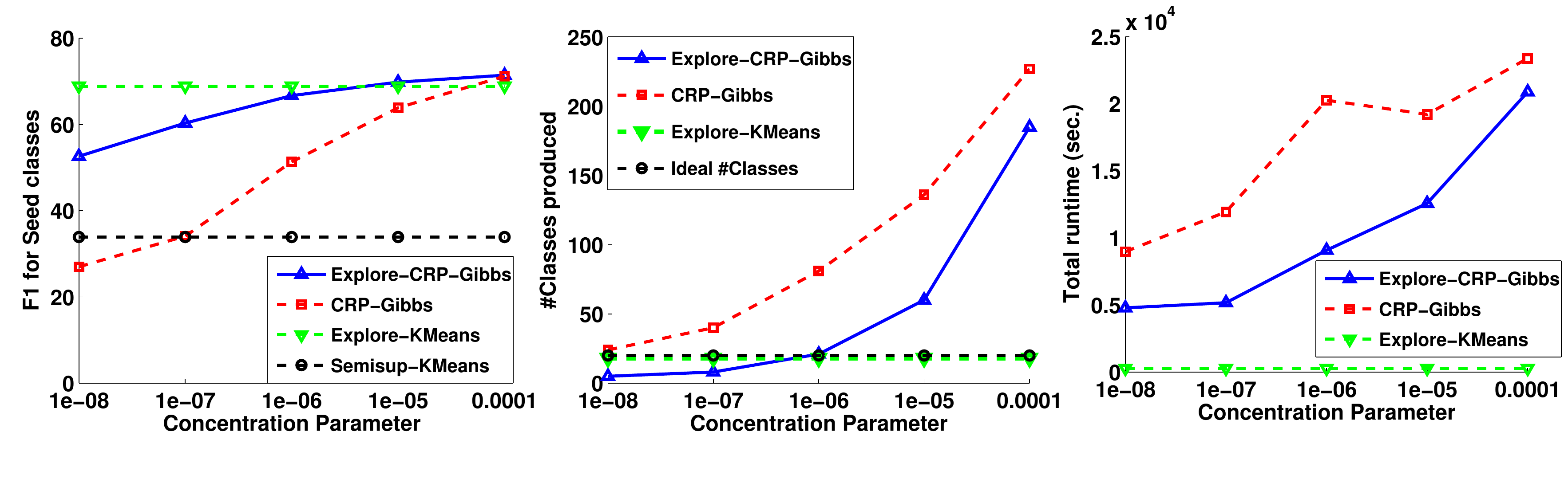}
\vspace{-0.15in}
\caption{20-Newsgroups dataset: Varying concentration parameter, with six seed classes.}
\label{fig:20NGVaryPrior}
\end{figure*}

It is well-known that CRP is sensitive to the concentration
parameter $P_{\it new}$.  Figures \ref{fig:DelSpVaryPrior} and
\ref{fig:20NGVaryPrior} show the performance of all the exploratory
methods, as well as \semisupKM, as the concentration parameter is
varied from $10^{-8}$ to $10^{-2}$. (For \explKM\ and \semisupKM\
methods, this parameter is irrelevant).  We show F1, the number of
classes produced, and run-time (which is closely related to the number
of classes produced.)  The results show that a well-tuned seeded
CRP-Gibbs can obtain good F1-performance, but at the cost of
introducing many unnecessary clusters.  
The modified Explore-CRP-Gibbs
performs consistently better, but not better than \explKM{}, and
\semisupKM\ performs the worst. 

\section{Related Work} \label{sect:rel_work}
\vspace{-0.1in}
In this paper we describe and evaluate a novel multiclass SSL method
that is more robust when there are unanticipated classes in the
data---or equivalently, when the algorithm is given seeds from only
some of the classes present in the data.  To the best of our knowledge this
specific problem has not been explored in detail before, even though
in real-world settings, there can be unanticipated (and hence
unseeded) classes in any sufficiently large-scale multiclass SSL task.

More generally, however, it has been noted before that SSL may suffer
due to the presence of unexpected structure in the data. For instance,
Nigam et al's early work on SSL based EM with multinomial
Naive Bayes \cite{nigam:mlj2000} noted that adding too much unlabeled
data sometimes hurt performance on SSL tasks, and discusses several
reasons this might occur, including the possibility that there might
not be a one-to-one correspondence between the natural mixture
components (clusters) and the classes.  To address this problem, they
considered modeling the positive class with one component, and the
negative class with a mixture of components.  They propose to choose
the number of such components by cross-validation; however, this
approach is relatively expensive, and inappropriate when there are
only a small number of labeled examples (which is a typical case in
SSL).  More recently, McIntosh
\cite{McIntosh:2010:UDN:1870658.1870693} described heuristics for
introducing new ``negative categories'' in lexicon bootstrapping,
based on a domain-specific heuristic for detecting semantic drift with
distributional similarity metrics.  Our setting is broadly similar to
these works, except that we consider this task in a general
multiclass-learning setting, and do not assume seeds from an
explicitly-labeled ``negative'' class, which is a mixture; instead, we
assume seeds from known classes only.  Thus we assume that data fits a
mixture model with a one-to-one correspondence with the classes, but
only after the learner introduces new classes hidden in the data.
We also explore this issue in much more depth experimentally, by
systematically considering the impact of having too few seed
classes, and propose and evaluate a solution to the problem.
There has also been substantial work in the past to automatically
decide the right ``number of clusters'' in unsupervised learning
\cite{dutta:flairs2011,pelleg:icml2000,Hamerly:nips03,Chiang:jclass2010,Menasce:ec1999,welling:sdm2006}.  Many of these
techniques are built around \km{} and involve running it
multiple times for different values of K.  Exploratory learning differs in that we focus on a SSL
setting, and evaluate specifically the performance difference on the
seeded classes, rather than overall performance differences.

There is also a substantial body of work on constrained clustering;
for instance, Wagstaff et al \cite{Wagstaff:icml2001} describe a
constrained clustering variant of \km{} ``must-link'' and
``cannot-link'' constraints between pairs. This technique changes the
cluster assignment phase of \km{} algorithm by assigning each
example to the closest cluster such that none of the constraints are
violated.  SSL in general can be viewed as a special case of
constrained clustering, as seed labels can be viewed as constraints on
the clusters; hence exploratory learning can be viewed as a subtype of
constrained clustering, as well as a generalization of SSL.
However, our approach is different in the sense that 
there are more efficient methods for
dealing with seeds than arbitrary constraints.

In this paper we focused on EM-like SSL methods.  Another widely-used approach
to SSL is label propagation.  In the modified adsorption algorithm
\cite{Talukdar:ecml2009}, one such graph-based label propagation
method, each datapoint can be marked with one or more known labels, or
a special dummy label meaning ``none of the above''.  Exploratory
learning is an extension that applies to a different class of SSL
methods, and has some advantages over label propagation: for instance,
it can be used for inductive tasks, not only transductive tasks.
\explEM{} also provides more information by introducing multiple
``dummy labels'' which describe multiple new classes in the data.

A third approach to SSL involves unsupervised dimensionality reduction
followed by supervised learning (e.g., \cite{Dalvi:sdm2013}).
Although we have not explored their combination, these techniques are
potentially complementary with exploratory learning, as one could also
apply EM-like methods, in a lower-dimensional space (as is typically
done in spectral clustering).  If this approach were followed then an
exploratory learning method like \explEM{} could be used to introduce
new classes, and potentially gain better performance, in a
semi-supervised setting.

One of our benchmark tasks, entity classification, is inspired by the
NELL (Never Ending Language Learning) system \cite{Carlson:wsdm2010}.
NELL performs broad-scale multiclass SSL.  One subproject within NELL
\cite{Mohamed:emnlp2011} uses a clustering technique for discovering
new relations between existing noun categories---relations not defined
by the existing hand-defined ontology.  Exploratory learning addresses
the same problem, but integrates the introduction of new classes into
the SSL process.  Another line of research considers the problem of
``open information extraction'', in which no classes or seeds are
used at all
\cite{Textrunner:naacl2007,Etzioni:www2004,Dalvi:wsdm2012}.
Exploratory learning, in contrast, can exploit existing information
about classes of interest and seed labels to improve performance.

Another related area of research is novelty detection.
Topic detection and tracking task aims to detect novel documents at 
time $t$ by comparing them to all documents till time $t-1$ and 
detects novel topics. Kasiviswanathan et al. \cite{kasiviswanathan:cikm2011} assumes the number 
of novel topics is given as input to the algorithm. 
Masud et al. \cite{masud:ecml2009} develop techniques on streaming data to predict 
whether next data chunk is novel or not. Our focus is on improving performance 
of semi-supervised learning when the number of new classes is unknown. 
Bouveyron \cite{bouveyron:hal2010} worked on the EM approach to model unknown classes, but the entire 
EM algorithm is run for multiple numbers of classes. 
Our algorithm jointly learns labels as well as new classes. 
Sch{\"o}lkopf et al. \cite{scholkopf:NIPS2000} defines a problem of learning a function over the data 
space that isolates outliers from class instances. Our approach is different 
in the sense we do not focus on detecting outliers for each class.


\section{Conclusion} \label{sect:conclusions}
\vspace{-0.1in}
In this paper, we investigate and improve the robustness of SSL
methods in a setting in which seeds are available for only a subset of
the classes---the subset of most interest to the end user.  We
performed systematic experiments on fully-labeled multiclass problems,
in which the number of classes is known.  We showed that if a user
provides seeds for only some, but not all, classes, then SSL
performance is degraded for several popular EM-like SSL methods
(semi-supervised multinomial Naive Bayes, seeded \km, and a seeded
version of mixtures of von Mises-Fisher distributions).  We then
described a novel extension of the EM framework called \explEM{},
which makes these methods much more robust to unseeded classes.
Exploratory EM introduces new classes on-the-fly during learning based
on the intuition that hard-to-classify examples---specifically,
examples with a nearly-uniform posterior class distribution---should
be assigned to new classes.  The exploratory versions of these SSL
methods often obtained dramatically better performance---e.g.,
on Delicious\_Sports dataset up to 90\% improvements in F1,
on 20-Newsgroups dataset up to 27\% improvements in F1,
and on Reuters dataset up to 200\% improvements in F1.
In comparative experiments, one exploratory SSL method, \explKM, emerged as a strong
baseline approach.

Because \explEM{} is broadly similar to non-parametric Bayesian
approaches, we also compared \explKM{} to a seeded version of an
unsupervised mixture learner that explores differing numbers of
mixture components with the Chinese Restaurant process (CRP).
\explKM{} is faster than this approach, and more accurate as well,
unless the parameters of the CRP are very carefully tuned. \explKM{}
also generates a model that is more compact, having close to the true
number of clusters.  The seeded CRP process can be improved, moreover,
by adapting some of the intuitions of \explKM, in particular by
introducing new clusters most frequently for hard-to-classify
examples (those with nearly-uniform posteriors).

The exploratory learning techniques we described here are limited to
problems where each data point belongs to only one class.  An
interesting direction for future research can be to develop such
techniques for multi-label classification, and hierarchical
classification. Another direction can be create more scalable parallel
versions of Explore-KMeans for much
larger datasets, e.g., large-scale entity-clustering task.

\subsubsection*{Acknowledgments:}
This work is supported in part by the Intelligence Advanced Research Projects Activity
(IARPA) via Air Force Research Laboratory (AFRL) contract number
FA8650-10-C-7058. The U.S. Government is authorized to reproduce and
distribute reprints for Governmental purposes notwithstanding any
copyright annotation thereon. This work is also partially supported by the Google Research Grant.
The views and conclusions contained herein are those of the authors and should 
not be interpreted as necessarily representing the official policies or 
endorsements, either expressed or implied, of Google, IARPA, AFRL, or the U.S. Government. 

{
\small
\bibliography{references}
\bibliographystyle{abbrv}
}

\end{document}